\documentclass[10pt,runningheads]{llncs}
\usepackage{amsfonts}
\usepackage{graphicx}
\usepackage{subcaption}
\usepackage{tikz}
\usetikzlibrary{arrows,decorations.pathmorphing,positioning,backgrounds,automata}
\usepackage{pgf}
\usepackage{pgfplots}
\usepackage{multirow}
\usepackage{amsmath,amssymb}
\usetikzlibrary{arrows,decorations.pathmorphing,positioning,backgrounds}
\definecolor{olivegreen}{rgb}{0,0.6,0}
\pgfplotsset{compat=1.14}

\definecolor{ballblue}{rgb}{0.13, 0.67, 0.8}

\newcommand{\xx}{\mathbf{x}}
\newcommand{\cc}{\mathbf{c}}

\newcommand{\zz}{\mathbf{z}}
\newcommand{\rr}{\mathbf{r}}

\newcommand{\KL}{\textrm{KL}}

\begin{document}
\title{Conditioned Variational Autoencoder for top-N item recommendation}
%
%
\author{Tommaso Carraro\inst{1}\orcidID{0000-0002-3043-1456} \and
Mirko Polato\inst{1}\orcidID{0000-0003-4890-5020} \and
Fabio Aiolli\inst{1}\orcidID{0000-0002-5823-7540}}
\authorrunning{T. Carraro et al.}
%
\institute{Department of Mathematics, University of Padova, Padova, Italy}
\maketitle              
\begin{abstract}

In this paper, we propose a Conditioned Variational Autoencoder (C-VAE) for constrained top-N item recommendation where the recommended items must satisfy a given condition. The proposed model architecture is similar to a standard VAE in which the condition vector is fed into the encoder. The constrained ranking is learned during training thanks to a new reconstruction loss that takes the input condition into account. We show that our model generalizes the state-of-the-art Mult-VAE collaborative filtering model. Moreover, we provide insights on what C-VAE learns in the latent space, providing a human-friendly interpretation. Experimental results underline the potential of C-VAE in providing accurate recommendations under constraints. Finally, the performed analyses suggest that C-VAE can be used in other recommendation scenarios, such as context-aware recommendation.

\end{abstract}
\section{Introduction}

Recommender system (RS) technologies are nowadays an essential component for e-services. 
Generally speaking, an RS aims at providing suggestions for items (e.g., movies, songs, news) that are most likely of interest to a particular user~\cite{RSH2015}. Since the first appearance of RSs, Collaborative Filtering (CF)~\cite{Su2009,Koren2011} has affirmed of being the \textit{de facto} recommendation approach. CF exploits similarity patterns across users and items to provide personalized recommendations. Latent Factor models, in particular Matrix Factorization (MF), have dominated the CF scene~\cite{Hu2008,Rendle2010,Ning2011,Polato2018} for years, and this has been further emphasized with the deep learning rise~\cite{Goodfellow2016}. 
A growing body of work has shown the potential of (deep) neural network approaches to face the recommendation problem. In the last few years, plenty of neural network-based models have raised the bar in terms of recommendation accuracy, such as NeuCF~\cite{He2017}, CDAE~\cite{Wu2016}, and EASE~\cite{Steck2019}, to name a few.

Recently, generative approaches have attracted the researchers' attention to the top-N recommendation task. The first generative models that appeared in the RS literature were based on Generative Adversarial Networks~\cite{Wang2017,Kang2017,Wang2018,Chae2018}. The GAN-based trend has been followed by a series of Variational Autoencoder-based (VAE, \cite{KingmaW13}) methods, which have soon gained much success overshadowing GAN-based ones. The seminal Variational approach for CF has been Mult-VAE~\cite{Liang2018}.
After that, other VAE-based models have been proposed, such as \cite{Li2018} and \cite{Iqbal2019} for the content-based recommendation. In particular, the latter work is highly related to ours since it extends the Conditional VAE~\cite{Zhang2016,Pagnoni2018ConditionalVA} to collaborative filtering (discussed in Section~\ref{sec:scr}).

Here, we extend the Mult-VAE model~\cite{Liang2018} for conditioned recommendations in the top-N setting. Conditions are intended in a generic sense and they can be both content-based or contextual-based~\cite{Adomavicius2015}.
It is important to underline that the way conditions are treated in our training is different from the one proposed in~\cite{Iqbal2019}. In our setting, a condition represents a constraint, and thus the provided recommendations must satisfy the constraint to be accepted. For example, in a movie recommendation system a user can ask for movies that belong to a specific genre. 

We designed our model as a generalization of Mult-VAE, and hence if trained without the conditions they are equivalent. Treating the conditions as we do allows the model to be versatile making it potentially applicable as a content-based as well as a context-aware recommender. 
Additionally, thanks to the training process, the latent space shows nice properties that can be exploited to give a human-friendly interpretation of the model.
Our experimental analyses show that our method can achieve state-of-the-art performance on different benchmark data sets. 

In summary, 
our main contributions are:
\begin{enumerate}
    \item we propose a Conditioned VAE for top-N recommendation, dubbed C-VAE, able to manage conditioned recommendations. We define the C-VAE architecture and a new conditioned loss, with which our model is able to learn the relationships between items and conditions;
    \item we provide a descriptive as well as a quantitative comparison with state-of-the-art approaches;
    \item we provide a in-depth analysis of the properties of the learned latent space giving a human-friendly interpretation.
\end{enumerate}

The remainder of the paper is structured as follows. Section~\ref{sec:background} provides the background useful to follow the rest of the paper. Section~\ref{sec:cvae} describes the proposed method, while Section~\ref{sec:exp} shows the performed evaluation. Finally, Section~\ref{sec:conc} wraps up the paper and gives some possible future research paths.

\section{Background}\label{sec:background}
This section provides the notations (Section~\ref{sec:notation}) and background knowledge  (Section~\ref{sec:vae}) useful to fully understand the rest of the paper.

\subsection{Notation}\label{sec:notation}
In this section we provide some useful notation used throughout the paper. We refer to the set of users of a RS with $\mathcal{U}$, where $|\mathcal{U}|=n$. Similarly, the set of items is referred to as  $\mathcal{I}$ such that $|\mathcal{I}|=m$. The set of ratings is denoted by $\mathcal{R} \equiv \{(u,i) \mid u \in \mathcal{U} \wedge i \in \mathcal{I}, u \textrm{ rated } i \}$. Each item $i$ is assumed to belong to a set of categories $C^{(i)} \subseteq \mathcal{C}$, where $\mathcal{C} \equiv {\{C_1, C_2, \dots, C_s\}}$ is the set of all possible categories s.t. $|\mathcal{C}|=s$.

Moreover, since we face top-N recommendation tasks, we consider the binary rating matrix with $\mathbf{R} \in \mathbb{R}^{n \times m}$, where users are on the rows and items on the columns, such that $\mathbf{r}_{ui}=1$ iff $(u,i) \in \mathcal{R}$. Given $\mathbf{R}$, $\mathbf{r}_{u} \{0,1\}^m$ indicates the column binary vector corresponding to the user $u$.
We add a subscription to both user and item sets to indicate, respectively, the set of items rated by a user $u$ (i.e., $\mathcal{I}_u$) and the set of users who rated the item $i$ (i.e., $\mathcal{U}_i$). 
Finally, we indicate with $\mathbf{c}=[{c}_{1},...,{c}_{s}]^\top$ the column binary condition vector, where $c_j = 1$ if and only if $\exists i \in \mathcal{I}_u$ such that  $i$ belongs to the category $C_j$. 

\subsection{Variational Autoencoder}\label{sec:vae}
The VAE is a generative model that assumes the input $\xx$ is generated according to the following generative process: $\zz \sim p_{\theta^*}(\zz)$ and $\xx|\zz \sim p_{\theta^*}(\xx|\zz)$, where the dimensionality of $\zz$ is (generally) much lower than $\xx$. 
In other words, VAE assumes that the input vector $\xx$ is modeled as a function of an unobserved random vector $\zz$ of lower dimensionality. 
VAE aims at estimating the parameters $\theta^*$ by maximizing the likelihood of the data (Maximum Likelihood Estimation, MLE), i.e., $\hat{\theta}=\arg \max_{\theta \in \Theta}  \: p_{\theta}(\xx)$.
Computing the MLE requires solving
$$ p_{\theta}\left(\xx\right)=\int p_{\theta}(\xx | \zz) p_{\theta}(\zz) d \zz $$
which is often intractable. However, in practice, for most $\zz$, $p_{\theta}(\xx|\zz) \approx 0$. The key idea behind VAE is to sample values of $\zz$ that are likely to have produced $\xx$, and compute $p_\theta(\xx)$ just from those. To do this, we need an approximation $q_\phi(\zz|\xx)$ of the true posterior distribution that returns a distribution over $\zz$ that are likely to produce the input. To make this problem tractable, it is assumed that $q_\phi$ follows a specific family of parametric distributions, usually a normal distribution with 0 mean and unitary variance. The closeness between $q_\phi(\zz|\xx)$ and the assumed posterior distribution $p_\theta(\zz|\xx)$ is ensured by the minimization of the Kullback-Liebler divergence (KL), which can be written as:
\begin{equation}\label{eq:kl}
\KL(q_\phi(\zz | \xx) \| p_\theta(\zz | \xx))=\mathbb{E}_{q_{\phi}(\zz | \xx)}\left[\log q_\phi(\zz | \xx) - \log p_\theta(\xx, \zz)\right]+\log p_\theta(\xx).
\end{equation}

After some rearrangements of Equation~\eqref{eq:kl} it is possible to write the so-called Evidence Lower BOund (ELBO)~\cite{KingmaW13}, which naturally defines the objective function that VAE wants to maximize:
$$
\begin{aligned}
\log p_\theta(\xx) & \geq \mathbb{E}_{q_{\phi}(\zz | \xx)}\left[\log p_{\theta}(\xx | \zz)\right] - \KL (q_{\phi}(\zz | \xx) \| p_\theta(\zz)) \\
& = \mathcal{L}\left(\xx; \theta, \phi\right).
\end{aligned}
$$
This loss can be interpreted as a reconstruction loss (first term), plus the so-called KL loss which acts as a kind of regularization term.

In practice, $p_\theta$ and $q_\phi$ are parametrized by two (deep) neural networks, i.e., the decoder ($f_\theta$) and the encoder ($g_\phi$), respectively. These parameters are optimized using stochastic gradient ascent with the aid of the reparameterization trick~\cite{KingmaW13}, that allows to compute the gradient w.r.t $\phi$.
To this end, the encoder network provides the parameters which define the distributions over each element of $\zz$, i.e, mean $\boldsymbol{\mu}$ and variance $\boldsymbol{\sigma}$. The sampling over the Gaussian distribution is performed via an additional input $\boldsymbol{\epsilon}$, which allows the reparameterization 
$\zz = \boldsymbol{\mu} + \boldsymbol{\epsilon} \odot \boldsymbol{\sigma}$, where $\odot$ is the Hadamard product.

\subsection{Variational Autoencoder for collaborative filtering}\label{sec:vaecf}
In~\cite{Liang2018}, Liang et al. propose a VAE for collaborative filtering called Mult-VAE. Mult-VAE takes as input the user-item binary rating matrix and learns a compressed latent representation (the encoder) of the input. These latent representation is then used to reconstruct the input (the decoder) and to impute the missing ratings. The top-N recommendation is computed by taking, for each user, the N items with the highest reconstructed ratings. 

Differently from the standard VAE, Mult-VAE uses a multinomial log likelihood rather than the classical Gaussian likelihood. Authors believe that the multinomial distribution is well suited for modeling implicit feedback~\cite{Liang2018}. Moreover, Mult-VAE employs a $\beta$-VAE loss~\cite{Higgins2017betaVAELB} in which the hyper-parameter $\beta$ is added to the loss as a trade-off parameter between the reconstruction loss and the KL loss. 

\subsection{Style Conditioned Recommendation (SCR)}\label{sec:scr}
In \cite{Iqbal2019} a style conditioned variational autoencoder is proposed. The conditional schema followed by SCR is similar to the standard Conditional VAE~\cite{Zhang2016,Pagnoni2018ConditionalVA}. The style conditioning is achieved with the addition of a user style profile vector to both the input of the encoder and the decoder. This style vector representation is learned though another network using side information. The rest of the model as well as the training of the network is the same as in a standard VAE, where the input of the encoder and the decoder is the concatenation of their input with the style vector. 

\section{Conditioned Variational Autoencoder}\label{sec:cvae}
In this section we present C-VAE for top-N recommendation. We first underline the differences with the state-of-the-art (Section~\ref{sec:preliminaries}), and then we define the architecture (Section~\ref{sec:arc}) as well as the new loss (Section~\ref{sec:loss}) of our model.

\subsection{Preliminaries}\label{sec:preliminaries}
Once learned the user style profile, the conditioning proposed in~\cite{Iqbal2019} is fixed for the user. The only way to force different styles is by acting directly in the latent space by injecting a specific style via a one-hot encoded style vector. With this trick the decoder network is driven to reproduce inputs with a specific style. However, this is a post-training step, and hence only the decoding is influenced by the injection. 

Since our conditions represent constraints, in C-VAE the conditional vector is fed into the encoder network, and the training process guarantees that the learned latent representation depends on the given condition. In this way, different conditions map the user onto (potentially) different regions of the latent space. In Section~\ref{sec:latent} we show that this is a nice property when it comes to interpret the model.
C-VAE differs from SCR in the following main aspects:
\begin{itemize}
    \item C-VAE is architecturally simpler than SCR (Section~\ref{sec:arc});
    \item C-VAE is more versatile than SCR and it is a natural generalization of the Mult-VAE (Section~\ref{sec:vaecf}) model;
    \item C-VAE learns the correlations between users and conditions (SCR only learns a specific conditioning for each user).
\end{itemize}

\subsection{Architecture}\label{sec:arc}

C-VAE follows the architecture of Mult-VAE with the addition of a conditional vector to the input of the encoder network. The conditional vector is concatenated with the user rating vector after the dropout layer. The dropout layer (implemented in \cite{Liang2018} but not reported in the paper) gives to the VAE denoising capabilities, which have shown of being effective in making recommendations~\cite{Liang2018}. The conditional vector $\cc \in \{0,1\}^s$ is defined as a one-hot vector over the $s$ possible conditions. It is noteworthy that the condition, differently from SCR, is not fed into the decoder network.

\begin{figure}[htbp]
    \centering
    \includegraphics{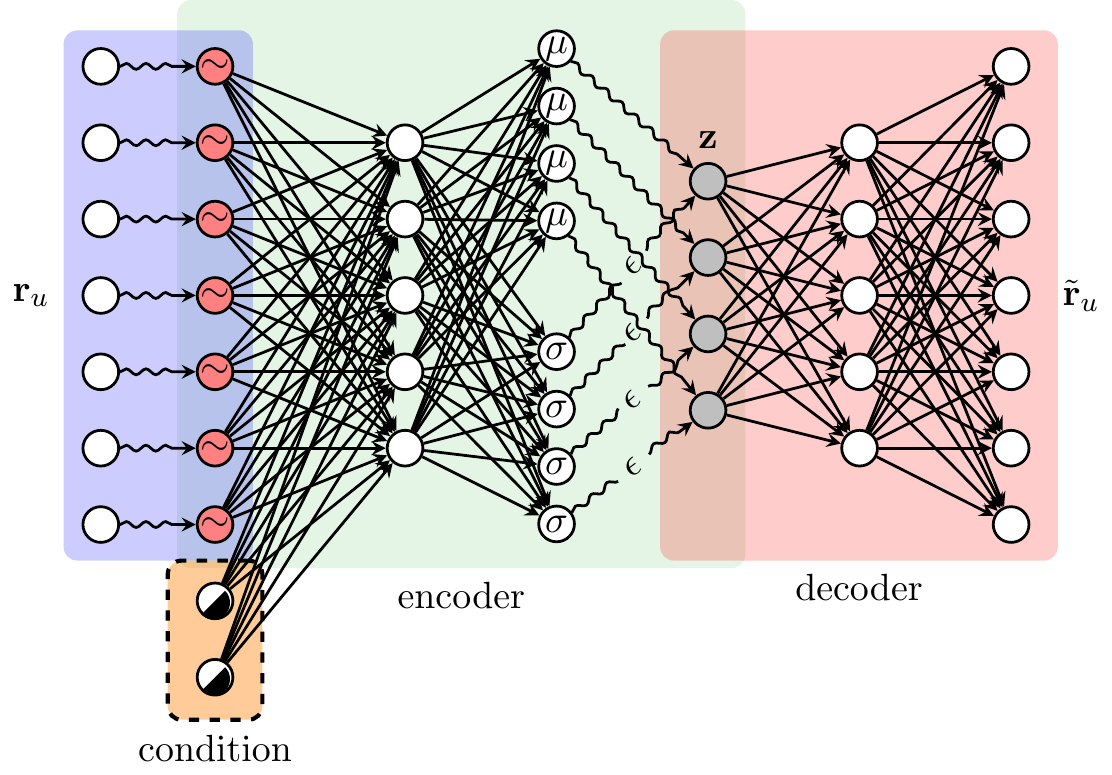}
    \caption{High level illustration of the Conditioned VAE architecture.}
    \label{fig:cvae-archi}
\end{figure}

\begin{figure*}[h]
\begin{subfigure}{.27\textwidth}
\centering
\begin{tikzpicture}[->,>=stealth',shorten >=1pt,auto,node distance=1.8cm,
                    semithick]
  \node[state] (mu)                    {$\mu$};
  \node[state,fill=lightgray]         (x) [above right of=mu] {$\xx$};
  \node[state]         (z) [below right of=mu] {$\zz$};
  \node[state]         (std) [above right of=z] {$\sigma$};
  \node[state]         (o) [below of=z]       {$\tilde{\xx}$};
  \node[state, fill=lightgray]         (eps) [right of=z]       {$\epsilon$};

  \path (mu) edge              node {} (z)
		 (x) edge node[fill=white, anchor=center, pos=0.4,font=\footnotesize] {$\phi$} (mu)
       		edge node[fill=white, anchor=center, pos=0.4,font=\footnotesize] {$\phi$} (std)
        (std) edge              node {} (z)
        (z) edge node[fill=white, anchor=center, pos=0.4,font=\footnotesize] {$\theta$} (o)
        (eps) edge[dashed] node {} (z);
\end{tikzpicture}
\caption{Mult-VAE}
\label{fig:vae}
\end{subfigure}
\begin{subfigure}{.34\textwidth}
\centering
\begin{tikzpicture}[->,>=stealth',shorten >=1pt,auto,node distance=1.8cm,
                    semithick]
  \node[state] (mu)                    {$\mu$};
  \node[state,fill=lightgray]         (x) [above right of=mu] {$\xx$};
  \node[state]         (z) [below right of=mu] {$\zz$};
  \node[state]         (std) [below right of=x] {$\sigma$};
  \node[state]         (o) [below of=z]       {$\tilde{\xx}$};
  \node[state, fill=lightgray]         (c) [left of=x]       {$\cc$};
  \node[state, fill=lightgray]         (eps) [right of=z]       {$\epsilon$};

  \path (mu) edge              node {} (z)
		 (x) edge node[fill=white, anchor=center, pos=0.4,font=\footnotesize] {$\phi$} (mu)
       		edge node[fill=white, anchor=center, pos=0.4,font=\footnotesize] {$\phi$} (std)
        (std) edge              node {} (z)
        (z) edge node[fill=white, anchor=center, pos=0.4,font=\footnotesize] {$\theta$} (o)
        (c) edge[dotted] node {} (x)
            edge[dotted,bend right=60] node {} (z)
        (eps) edge[dashed] node {} (z);
\end{tikzpicture}
\caption{SCR}
 \label{fig:scr}
\end{subfigure}
\begin{subfigure}{.3\textwidth}
\centering
\begin{tikzpicture}[->,>=stealth',shorten >=1pt,auto,node distance=1.8cm,
                    semithick]
  \node[state] (mu)                    {$\mu$};
  \node[state,fill=lightgray]         (x) [above right of=mu] {$\xx$};
  \node[state]         (z) [below right of=mu] {$\zz$};
  \node[state]         (std) [below right of=x] {$\sigma$};
  \node[state]         (o) [below of=z]       {$\tilde{\xx}$};
  \node[state, fill=lightgray]         (c) [left of=x]       {$\cc$};
  \node[state, fill=lightgray]         (eps) [right of=z]       {$\epsilon$};

  \path (mu) edge              node {} (z)
		 (x) edge node[fill=white, anchor=center, pos=0.4,font=\footnotesize] {$\phi$} (mu)
       		edge node[fill=white, anchor=center, pos=0.4,font=\footnotesize] {$\phi$} (std)
        (std) edge              node {} (z)
        (z) edge node[fill=white, anchor=center, pos=0.4,font=\footnotesize] {$\theta$} (o)
        (c) edge[dotted] node {} (x)
        (eps) edge[dashed] node {} (z);
\end{tikzpicture}
\caption{C-VAE}
 \label{fig:cvae}
\end{subfigure}
\caption{Overview of the architectural differences between (a) Mult-VAE, (b) SCR, and (c) C-VAE. The dashed arrows denote a sampling operation, while the dotted arrows indicate the conditional input.}
\label{fig:architectures}
\end{figure*}
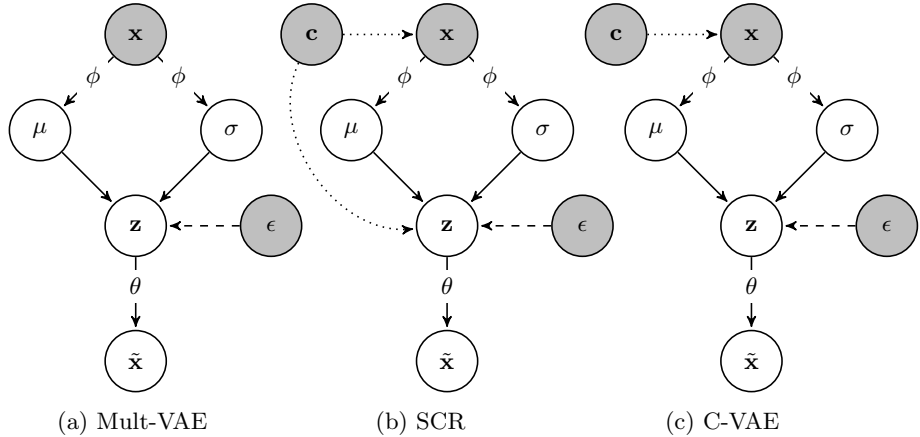

Figure~\ref{fig:cvae-archi} depicts the network architecture of our model, while Figure~\ref{fig:architectures} provides an overview of the architectural differences between Mult-VAE, SCR, and C-VAE.

\subsection{Conditioned loss function}\label{sec:loss}
A core difference between our C-VAE and SCR is the way the training works. Since we treat the conditioning as a constraint, the reconstruction must take this into account. This means that, in general, the expected output is a filtered version of the input, where the items that do not satisfy the constraint are dropped. This is achieved by our model via a modified loss function.

The loss function we try to minimize is a conditioned version of the Mult-VAE loss~\cite{Liang2018}:
$$
\mathcal{L}_{\beta}\left(\mathbf{r}_{u},\mathbf{c} ; \theta, \phi\right) = \mathbb{E}_{q_{\phi}\left(\mathbf{z}_{u} | \mathbf{r}_{u},\mathbf{c}\right)}\left[\log p_{\theta}\left(\hat{\mathbf{r}}_{u} | \mathbf{z}_{u},\mathbf{c}\right)\right]-\beta \cdot \textrm{KL}\left(q_{\phi}\left(\mathbf{z}_{u} | \mathbf{r}_{u},\mathbf{c}\right) \| p\left(\mathbf{z}_{u},\mathbf{c}\right)\right)
$$
where $\mathbf{z}_{u}$ is the latent representation of the user $u$, and $\hat{\mathbf{r}}_u$ is $\mathbf{r}_u$ filtered by the condition $\cc$. The filtering is directly embedded in the reconstruction loss as:
$$
\log p_{\theta}\left(\hat{\mathbf{r}}_{u} | \mathbf{z}_{u}, \mathbf{c}\right) = \sum\limits_{i \in \mathcal{I}}  \langle \mathbf{c}, \mathbf{G}^\top \rangle_i \:\mathbf{r}_{ui}  \log \pi_i(f_\theta(\zz_u)) 
$$
where:
\begin{itemize}
    \item $\pi$ is the softmax function;
    \item $\langle \cdot, \cdot \rangle$ indicates the inner-product operation;
    \item $\mathbf{G}\in\{0,1\}^{m\times s}$ is the item-condition matrix, where $g_{ic}=1$ iff item $i$ satisfies the condition $c$.
\end{itemize}
This reconstruction loss is what makes our model able to learn the relationship between items and conditions. 
When conditions are not used, then all items are assumed of satisfying the empty condition. Implementation-wise, this is achieved by removing the dot product $\langle \mathbf{c}, \mathbf{G}^\top \rangle$, which is indeed always equal to the constant vector $\mathbf{1}$. This leads the C-VAE loss to be the equal to the loss function defined in~\cite{Liang2018}, making C-VAE equivalent to Mult-VAE.

It is worth to underline that the filtering part can also be dependent from both user and item. For example, in the case of context-aware recommendation~\cite{Adomavicius2015}, the conditioning can be defined in terms of the context, which is influenced by both users and items. In this case, the reconstruction loss must be modified accordingly by defining a different condition matrix $\mathbf{G}$. In this paper, we focus on conditioning defined over the items' content.

\section{Experiments}\label{sec:exp}
In this section we present the experiments performed on C-VAE. We compared C-VAE with Mult-VAE in terms of top-N recommendation accuracy (Section~\ref{sec:res}). We simulate the conditioning on Mult-VAE by filtering its output according to the given condition. Finally, we analyze the latent space of C-VAE which allowed us to shed some lights on what happens under hood.

\subsection{Datasets}
We performed our experiments on three real-world data sets, chosen in such a way that they contained items side information to construct the conditions.
\begin{description}
    \item[] \texttt{MovieLens 20M}\footnote{\url{https://grouplens.org/datasets/movielens/20m/}} (\texttt{ml-20m}): This data set contains user-movie ratings collected from a movie recommendation service. We took the genres of the movies as conditions.
    We removed the rarest genres (i.e., \texttt{IMAX}, \texttt{Film-Noir} and the neutral genre \texttt{(no genres listed)}) because they were poorly represented in the data set. For the rest of the pre-processing we followed the same procedure as in~\cite{Liang2018}.
    
    \item[] \textbf{\texttt{Yelp}}\footnote{\url{https://www.yelp.com/dataset}}: This data set is a subset of Yelp's businesses, reviews, and user data. It was originally put together for the Yelp Data set Challenge. 
    We took the categories of the businesses as conditions. We kept the 20 most popular restaurant categories as described in~\cite{Spagnuolo_2020}.
    Afterwards, since we work on implicit feedback we binarized explicit data by keeping ratings of three or higher. Finally, we only kept users who have reviewed at least four restaurants and restaurants that have been reviewed by at least ten users.
    
    \item[] \texttt{Netflix Prize}\footnote{\url{https://www.kaggle.com/netflix-inc/netflix-prize-data}}: This is the official data set used in the Netflix Prize competition. As of \texttt{ml-20m}, we took the genres of the movies as conditions. Since the dataset does not include information about the genres, we developed a script to fetch these information from the \textit{IMDb}\footnote{\url{https://www.imdb.com/}} database. We used the title and year of the movies to query the database. Netflix dataset originally contained 17770 movies but only 12279 matched with \texttt{IMDb}. The retrieved movies-genres mapping is available at this URL\footnote{\url{https://github.com/bmxitalia/netflix-prize-with-genres}}.
    As in movielens, we removed the rarest genres, i.e., \texttt{Talk-Show}, \texttt{Film-Noir}, \texttt{Short}, \texttt{Reality-TV}, \texttt{News} and \texttt{Game-Show}. For the rest of the pre-processing we followed the same procedure as in~\cite{Liang2018}.
\end{description}

Table~\ref{tab:datasets} summarizes the information about the data sets after the pre-processing presented above.
    
\begin{table}[htbp]
\caption{Composition of datasets after pre-processing.}\label{tab:datasets}
\centering
\begin{tabular}{l c c c}
\hline
 &  \texttt{ml-20m} & \texttt{Yelp} & \texttt{Netflix}\\
\hline
\# of users & 136,466 & 125,679 & 459,133\\
\# of items & 19,619 & 22,824 & 11,844\\
\# of categories & 17 & 20 & 21\\
\# of interactions & 19.3M & 2.9M & 88.8M\\
\% of interactions & 0.7 & 0.1 & 1.6\\
\# of held-out users & 10,000 & 9,000 & 40,000\\
\hline
\# training examples & 1,728,205 & 759,955 &  6,826,774\\
\# validation examples & 144,179 & 47,364 &  699,901\\
\# test examples & 143,965 & 46,847 &  700,393\\
\hline
\end{tabular}
\end{table}
    
\subsection{Conditions computation}
Potentially, during the training of C-VAE it might be possible to condition each user with every possible conditions combination. However, the size of the training set would be in the order of $\mathcal{O}(n \cdot s^2)$. We decided to limit its size by conditioning users one category at a time, i.e., $\|\cc\|_1 = 1$. If a condition is never satisfied by the user's item set $\mathcal{I}_u$ then the condition is simply not considered in the training. In the training set we also considered the users without any condition (akin Mult-VAE).

The bottom part of Table~\ref{tab:datasets} summarizes the size of the training, validation and test sets after the computation of the conditions.

\subsection{Model architecture}
An overview of the architecture of our model is presented in Section~\ref{sec:arc}. We followed the implementation as in~\cite{Liang2018}, where an $L^2$ normalization and a dropout layer ($p=0.5$) are applied to the input $\rr_u$ before it is fed to the encoder. The encoder network is composed of a fully connected layer made of 600 neurons with \textit{tanh} as activation function. The encoder outputs the mean and the standard deviation of a Gaussian distribution, that are represented with two fully connected layers made of 200 neurons and linearly activated.
The decoder network is composed of a fully connected layer made of 600 neurons with \textit{tanh} as activation function. Finally, the decoder linearly outputs the scores over the entire items set.

Recalling that $m = |I|$ and $s = |\mathcal{C}|$, the neural architecture of our model can be summarized as $[m + s \implies 600 \implies 200 \implies 600 \implies m]$. For Mult-VAE we used the same architecture.

\subsection{Model training and hyper-parameters tuning}
For both Mult-VAE and C-VAE the network weights are initialized with Xavier uniform initializer, while biases are normally initialized with 0 mean and standard deviation 0.001. We used the Adam optimizer with learning rate 0.001. For the tuning of the hyper-parameter $\beta$ we used the procedure explained in~\cite{Liang2018}. As a reminder, this is the procedure we followed:
\begin{itemize}
    \item we trained the model annealing $\beta$ in such a way to reach 1.0 at the end of the training;
    \item we selected the $\beta$ value corresponding to the highest validation score in terms of nDCG@100~\cite{ndcg:2000};
    \item we re-trained the model annealing $\beta$ in such a way to reach the selected value at the end of the training.
\end{itemize}

In our experiments we found that the best values for $\beta$ are 0.07 for \texttt{ml-20m}, 0.35 for \texttt{Yelp} and 0.05 for \texttt{Netflix}.
We used a batch size of 500 for \texttt{Yelp} and \texttt{ml-20m}, while for \texttt{Netflix} a batch size of 1000. We trained the models for 100 epochs on every data set and we kept the model which corresponded to the best validation score. We used early stopping to stop the training if no improvements were found on the validation score for 5 consecutive epochs. 

\subsection{Experimental results and discussion}\label{sec:res}
In this section we compare C-VAE and Mult-VAE in terms of top-N reocmmendation quality. 
We used recall@k and nDCG@k as ranking-based metrics. While recall@k considers all items ranked within the first $k$ to be equally important, nDCG@k uses a monotonically increasing
discount to emphasize the importance of higher ranked items. We did not have the chance to compare our method with SCR because authors did not provide the implementation of their method. Experiments have been performed using the \textit{rectorch}\footnote{\url{https://github.com/makgyver/rectorch}} python library.
To quantitatively compare our proposed method with Mult-VAE we used three different types of evaluation:
\begin{enumerate}
    \item \textbf{total}: measures how well the model performs in general, that is when the test set contains both users with conditions and users without conditions;
    \item \textbf{normal}: measures how well the model performs without conditioning, that is when test set contains only users without conditions; 
    \item \textbf{conditioned}: measures how well the model performs with conditioning, that is when the test set contains only users with conditions.
\end{enumerate}
Since Mult-VAE does not directly handle the conditioning, we filtered its output according to the condition and we computed the ranking only on those items that satisfy the condition. It is worth to notice that this filtering gives Mult-VAE a huge advantage because it greatly narrows down the item set. Clearly, C-VAE instead performs the ranking over the whole item set.

To validate and test the models, for each validation/test user we fed 80\% of user ratings to the network and reported metrics on the remaining 20\% of the ratings history (for \texttt{Yelp} we used 50/50 proportions due to its sparsity). Table~\ref{table-results} reports the obtained results. 
\vspace{-1em}
\begin{table}[h!]
\caption{Comparison between C-VAE and Mult-VAE on selected benchmark data sets. Standard errors are between 0.001 and 0.003. r stands for recall, while n stands for nDCG. Each metric is averaged across all test users.}\label{table-results}
\centering
\begin{tabular}{l | l | c c c | c c c | c c c}
\hline
\multirow{2}{*}{Dataset} & \multirow{2}{*}{Method} & \multicolumn{3}{c |}{Total} & \multicolumn{3}{c |}{Normal} & \multicolumn{3}{c}{Conditioned}\\
&  & r@20 & r@50 & n@100 & r@20 & r@50 & n@100 & r@20 & r@50 & n@100 \\
\hline
\multirow{2}{*}{\texttt{ml-20m}} &
{C-VAE} & 0.638 & 0.786 & 0.509 & 0.385 & 0.527 & 0.410 & 0.666 & 0.816 & 0.521 \\
& {Mult-VAE} & \textbf{0.645} & \textbf{0.792} & \textbf{0.517} & \textbf{0.394} & \textbf{0.537} & \textbf{0.420} & \textbf{0.674} & \textbf{0.822} & \textbf{0.529} \\
\hline
\multirow{2}{*}{\texttt{Yelp}} &
{C-VAE} & \textbf{0.311} & 0.459 & \textbf{0.238} & \textbf{0.139} & \textbf{0.235} & \textbf{0.143} & 0.392 & 0.564 & \textbf{0.282} \\
& {Mult-VAE} & \textbf{0.311} & \textbf{0.460} & \textbf{0.238} & 0.134 & 0.233 & \textbf{0.143} & \textbf{0.394} & \textbf{0.567} & \textbf{0.282} \\
\hline
\multirow{2}{*}{\texttt{Netflix}} &
{C-VAE} & 0.590 & 0.751 & 0.494 & 0.335 & 0.444 & 0.374 & 0.613 & 0.778 & 0.504 \\

& {Mult-VAE} & \textbf{0.601} & \textbf{0.758} & \textbf{0.504} & \textbf{0.352} & \textbf{0.457} & \textbf{0.389} & \textbf{0.623} & \textbf{0.785} & \textbf{0.515} \\
\hline
\end{tabular}
\end{table}



From the table, it is possible to observe that C-VAE obtains state-of-the-art results, even though generally a bit lower than Mult-VAE. We want to emphasize one more time that C-VAE performs the ranking over all items, while Mult-VAE only on the subset of items satisfying the condition.
This shows that our method is able to learn the relationships between items and categories, since it is able to push the items that belong to the target category at the top of the ranking. 

\subsection{Analysis of the C-VAE produced rankings}
\begin{figure}[htbp]
\centering
\includegraphics[scale=.8]{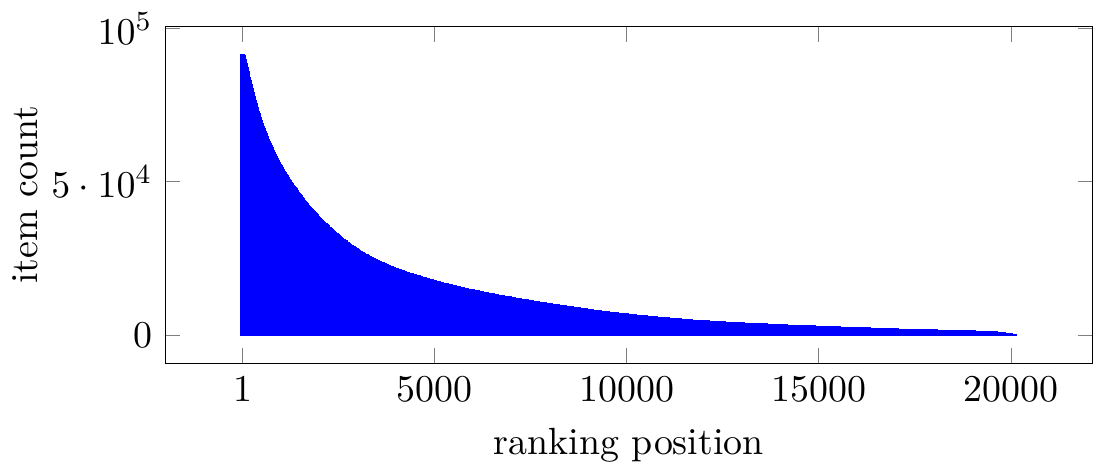}
\caption{Number of items of the target category per ranking position.}
    \label{fig-ranking}
\end{figure}

Since we obtained promising results in terms of ranking accuracy, we decided to analyze the categories distribution on the rankings produced by C-VAE. Thanks to the conditioned loss, C-VAE learns how to filter the items in such a way to focus its attention on the items belonging to the target category. 
To further validate this argument, we plot the distribution over the rankings produced by C-VAE for the items satisfying the conditions. The plot has been computed on the \texttt{ml-20m} training users and is shown in Figure~\ref{fig-ranking}.
It clearly shows that most of the items of the target category have been placed in the top positions, showing that C-VAE learns to push the right items at the top.
\vspace{-1em}

A further check have shown that in the first 100 positions of the ranking C-VAE is always able to put only items satisfying the input condition.

\subsection{Latent space exploration}\label{sec:latent}
Given the promising results discussed previously, we decided to further investigate the inner representations of C-VAE. In particular, to explore the latent space of the C-VAE model we took 2000 random users from the original \texttt{ml-20m} data set (no genres have been removed). We analyzed their learned latent representations by conditioning all of them on each genre, and also without the condition. We performed Principal Component Analysis (PCA)~\cite{Jolliffe2016PrincipalCA} and considered only the first 5 principal components. 

\begin{figure}[h]
\centering
\includegraphics[scale=.114]{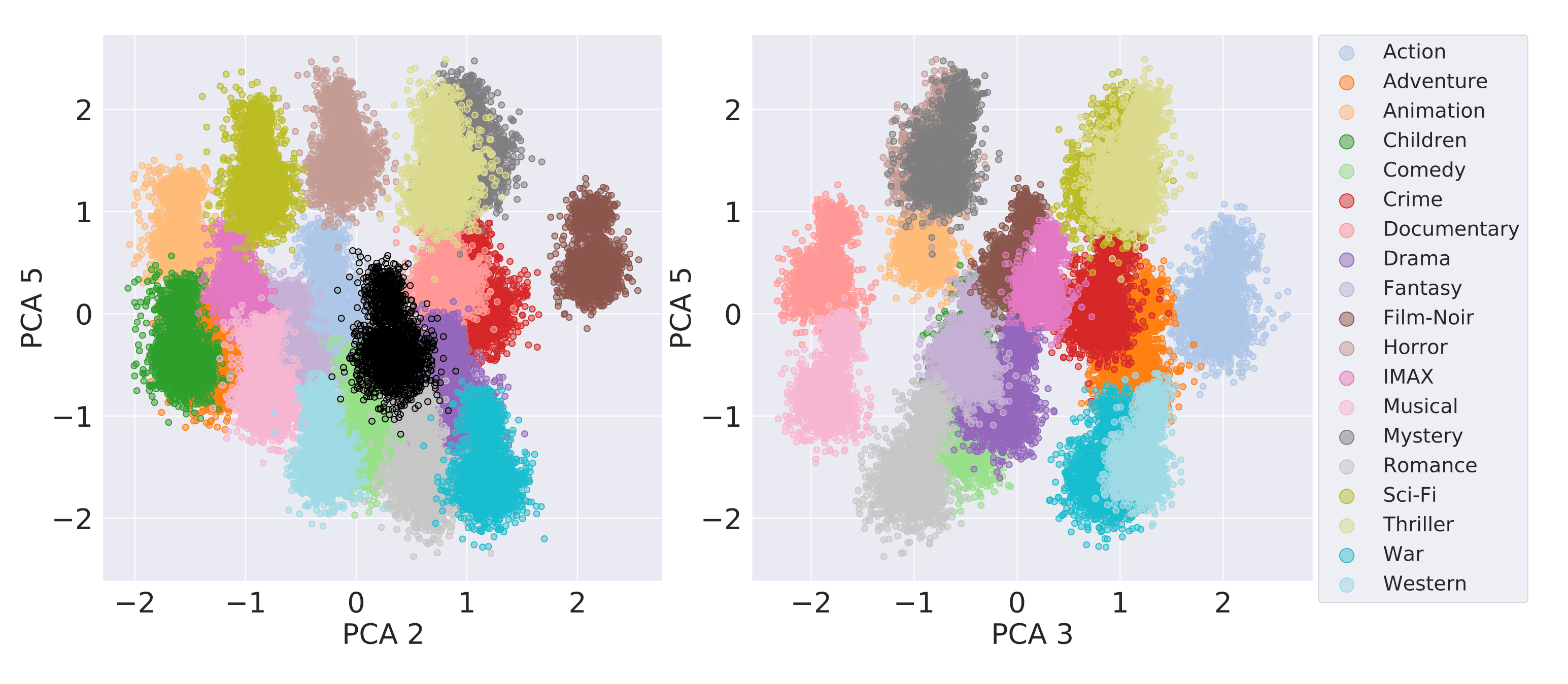}
\caption{(left) Second and fifth, (right) third and fifth components of PCA performed on selected users latent representations on \texttt{ml-20m}.\label{fig:pca}}
\end{figure}

We noticed that the first principal component separates the \texttt{(no genre listed)} genre from all the other genres. Thus, we decided to remove this \emph{neutral} genre and the first principal component. 
We also observed that the fourth principal component (not illustrated here) stretches the clusters on a single dimension, underling that even with the same conditioning users still have different tastes.
Principal components 2 and 5 showed really good properties giving an intuitive understanding of the genres correlation. Figure~\ref{fig:pca} (left hand side) plots these components. 

Looking at the figure, the following observations can be done:
\begin{itemize}
    \item popular and common genres (e.g., \texttt{Action}, \texttt{Comedy}, \texttt{Drama}, \texttt{Romance}) are placed close to the center of the latent space,
    while less popular ones (e.g., \texttt{Film-noir}, \texttt{Children}, \texttt{Animation}) are placed far aside;
    
    \item very different genres are placed distant from each other, while similar genres are placed near to each other. For example, \texttt{War} is distant from \texttt{Children} and \texttt{Animation}, while it is close to \texttt{Drama} and \texttt{Romance}; 
    
   
    \item the not conditioned representations (in black) are placed at the center of the space. We argue that when C-VAE recommends movies without the condition, it computes the unconditioned ranking and the most popular genres become more likely. 
\end{itemize}


The 2D plot of the third and fifth principal components (right hand side of Figure~\ref{fig:pca}) offers a different perspective with respect to the previous one. We argue that the third principal component captures the \emph{emotional theme} of the genres. For example, \texttt{Mistery} and \texttt{Horror} have similar emotional components (e.g., anxiety, tension, fear) and they almost completely overlap. Similar considerations can be done for \texttt{Children}-\texttt{Fantasy} and \texttt{War}-\texttt{Western}. 

\section{Conclusions and future work}\label{sec:conc}
In this paper, we presented a novel method for conditioning the top-N item recommendation process. We developed a conditioned extension of Mult-VAE~\cite{Liang2018}, which relies on a novel loss function that is crucial for the training of our method. 
We compared our method with the state-of-the-art for top-N recommendation, i.e., Mult-VAE, and we showed that C-VAE reaches similar performance in both the conditioned as well as the non conditioned top-N recommendation tasks.
Additionally, we explored the learned latent space of C-VAE and we observed that is able to capture not only the relationships between items and categories, but also the relationships between categories themselves. We also offered an intuitive and human-like interpretation of the latent representation.

As long as we perform content-based recommendations our model is less powerful than the filtered Mult-VAE, but in context-aware scenarios C-VAE can capture patterns between users, items and the interactions between them, while Mult-VAE cannot be applied since the filtering is no longer applicable.
In conclusion, it is our intent to extend our evaluation to other open research topics, in particular context-aware and also group recommender systems~\cite{Boratto2016}. 

%
%
%
%
%
%
\bibliographystyle{splncs04}
\bibliography{library}

\end{document}